%% file: main.tex
\documentclass[preprint,12pt]{elsarticle}

\usepackage{amssymb}
\usepackage{amsmath}
\usepackage{booktabs}
\usepackage{graphicx}
\usepackage{tikz}
\usetikzlibrary{positioning, arrows.meta, fit, backgrounds, calc}
\usepackage{xcolor}
\usepackage{hyperref}
\usepackage{multirow}
\usepackage{microtype}

\newcommand{\method}{SolCloudLLM}

\begin{document}

\begin{frontmatter}

\title{Bidirectional Multimodal Fusion of Sky Images and Time-Series for Solar Forecasting with Large Language Models}

\author[inst1]{Ken Chen\corref{cor1}}
\ead{kenchen1@student.unimelb.edu.au}
\author[inst1]{Maneesha Perera}
\author[inst1]{Wei Wang}
\author[inst1]{Sachith Seneviratne}
\author[inst2]{Hansani Weeratunge}
\author[inst1]{Saman Halgamuge}
\cortext[cor1]{Corresponding author.}
\affiliation[inst1]{organization={Department of Mechanical Engineering, The University of Melbourne},
    city={Melbourne},
    country={Australia}}
\affiliation[inst2]{organization={Department of Mechanical Engineering, Sri Lanka Institute of Information Technology},
    country={Sri Lanka}}

\begin{abstract}
Short-term photovoltaic (PV) power and global horizontal irradiance (GHI) forecasts are essential for effective dispatch, reserve scheduling, and grid operations with high PV penetration.
At these forecasting horizons, errors are predominantly driven by cloud induced ramps: relying solely on historical numerical data may struggle to anticipate an incoming cloud, making ground-based sky images a crucial complementary physical signal.
Furthermore, forecast performance is highly sensitive to location and local observing conditions, creating a strong need for site-specific data that are often scarce. 
Recently, large language models (LLMs) have demonstrated competitive performance and high data efficiency in time-series forecasting. Cross-modality alignment maps time-series data patches into the language embedding space, allowing pretrained representations to be leveraged for time-series forecasting with minimal task-specific data.
Despite their success, existing LLM-based forecasting methods remain predominantly unimodal, relying primarily on historical numerical time-series data. Effectively incorporating sky imagery into an LLM-based forecasting framework remains under-explored and an open challenge.
In this paper, we propose \method, an LLM-based multimodal forecasting framework.
\method~aligns sky-image patches with time-series patches and fuses their corresponding representations through bidirectional multimodal fusion, yielding a unified representation that is subsequently mapped into the embedding space of an LLM.
Extensive experiments on the SIRTA and SKIPP'D datasets demonstrate that \method~consistently outperforms the best baseline methods in Mean Squared Error (MSE) across all forecasting horizons ($H\in\{16,32,64\}$), achieving a maximum relative MSE reduction of $25.4\%$ on SIRTA.
Stratified analysis on SKIPP'D further indicates that the benefits of multimodal fusion are concentrated primarily under cloudy conditions at $H{=}32$ and $H{=}64$.
Notably, \method~achieves the best performance in nearly all few-shot settings (using only $5\%$ to $10\%$ of training data), whereas other deep learning baselines experience substantial performance degradation and are frequently outperformed by the non-learning physical method.
\end{abstract}


\end{frontmatter}


\input{sections/introduction}
\input{sections/related_work}
\input{sections/method}
\input{sections/experiments}

\input{sections/conclusion}

\bibliographystyle{elsarticle-num}
\bibliography{references}

\end{document}

%% file: sections/introduction.tex
\section{Introduction}
\label{sec:intro}

The reliable integration of solar energy into modern power grids requires precise short-term (i.e. intra-hour) forecasts of photovoltaic (PV) output and global horizontal irradiance (GHI). Such accurate predictions are the cornerstone for optimal dispatch, dynamic reserve scheduling, and stable market operations~\cite{pedro2012assessment}. In these short-term horizons, the main source of forecasting error is caused by the rapid fluctuations in solar power caused by moving clouds. While a passing cloud can drastically reduce solar irradiance in under a minute, historical numerical data is inherently blind to such approaching meteorological shifts~\cite{lin2023intra}. As a result, purely time-series forecasting methods are fundamentally reactive, typically adjusting only after a sudden drop in solar power output has been observed. Incorporating ground-based sky imagery helps address this blind spot. By continuously monitoring cloud trajectories and optical textures, these sky images supply a forward-looking visual context that purely numerical sequences lack~\cite{paletta2023advances,nie2024opensource}.

Beyond temporal dynamics, cross-site discrepancies in camera hardware, local microclimates, and PV array specifications induce significant domain shifts, causing models optimized for one facility often struggle to generalize to a new site without localized retraining \cite{nie2024multilocation,paletta2024transfer}. Compounding this issue is the high acquisition cost of paired multimodal datasets, which leaves most target sites with strictly limited historical records. Under such constrained conditions, developing highly data-efficient methods is no longer merely advantageous, but also becomes a strict operational necessity.

Recently, large language models (LLMs) have been adapted for numerical time-series forecasting, leveraging the rich sequential priors encoded within frozen pretrained transformers. This is typically achieved through cross-modality alignment, which maps patched time-series tokens into the language embedding space, enabling the backbone to generate forecasts without requiring training from scratch on new datasets \cite{zhou2023onefitsall,jin2024timellm}. This paradigm not only achieves competitive accuracy on widely used long-term forecasting benchmarks (including ETT, Weather, Electricity (ECL), Traffic, and ILI \cite{wu2022timesnet}), but also enables highly data-efficient few-shot adaptation when labeled data are scarce, as the majority of LLM parameters remain frozen. For instance, Time-LLM—a representative reprogramming framework—aligns patched numerical tokens with a frozen language backbone via cross-attention and prefix prompting \cite{jin2024timellm}. While this framework, along with broader LLM applications \cite{chandana2026llms}, has been introduced to solar power prediction \cite{lin2025timellm,fan2025solarllm}, existing pipelines remain strictly unimodal, relying solely on historical time-series numerical sequences without integrating synchronized sky images.

On the other hand, traditional sky-image nowcasting demonstrates that modeling cloud kinematics can significantly reduce forecast lag and preempt abrupt irradiance fluctuations \cite{paletta2022eclipse,nie2024skygpt}. Similarly, satellite-based cloud-tracking pipelines enhance downstream distributed PV forecasts, particularly when visual feature extraction is optimized end-to-end against the final power prediction \cite{perera2026cloud}. Concurrently, another line of research fuses ground-based sky imagery with historical PV or GHI sequences using task-specific convolutional or attention-based architectures \cite{sun2019sunset,venugopal2019cnn,paletta2023omnivision}. While these studies definitively establish the complementary nature of visual and temporal data, they offer no mechanisms for integrating sky images into the aforementioned LLM frameworks. Consequently, effectively injecting sky imagery into an LLM architecture, while preserving its inherent data efficiency, remains under-explored and a critical open challenge.

To address this gap, we propose \method, a novel multimodal reprogramming framework that seamlessly integrates sky imagery with numerical time-series sequences. Unlike recent vision-language approaches that rely on heavy late-fusion mechanisms (fusing modalities only after they have been independently encoded)~\cite{lin2025pvvlm}, \method~integrates the visual and numerical streams before aligning the fused representation with the language embedding. Specifically, sky imagery frames are encoded via a lightweight CNN, and the resulting visual features are patched and temporally aligned with time-series patches.
Furthermore, rather than employing plain concatenation, which merely juxtaposes the two token streams without cross-modal interaction, we introduce a token-level Bidirectional Multimodal Fusion module inspired by feature-wise linear modulation (FiLM)~\cite{perez2018film}. This mechanism enables each modality to dynamically generate scale-and-shift parameters for the other: sky-image tokens modulate time-series patches, allowing the recent numerical history to be interpreted under the current cloud state; conversely, time-series tokens modulate image patches, emphasizing or suppressing visual cues based on recent numerical context.
The modulated streams are then fused via concatenation and subsequent projection, forming a unified representation that is reprogrammed and mapped into a frozen LLM backbone. By maintaining a shared temporal index between modalities prior to mapping, \method~attaches precise cloud context to the corresponding look-back segments. Importantly, relying on a lightweight fusion design alongside a frozen LLM backbone ensures that \method~retains the exceptional data-efficient adaptation capabilities of LLM-based time-series forecasters, maintaining optimal performance even under strict few-shot constraints.

The main contributions are as follows.
\begin{enumerate}
\item 
We propose \method, a multimodal forecasting framework based on LLM - that precisely aligns sky-image tokens with time-series patches and fuses them via token-level bidirectional multimodal fusion before mapping the unified representation into a frozen LLM backbone.

\item Extensive evaluations on open access solar datasets - SIRTA~\cite{haeffelin2005sirta} and SKIPP'D~\cite{nie2023skippd} demonstrate that \method~consistently outperforms baselines in terms of Mean Squared Error (MSE) across forecasting horizons $H\in\{16,32,64\}$. Notably, it achieves peak relative MSE of $25.4\%$ on SIRTA ($H{=}32$) and $14.5\%$ on SKIPP'D ($H{=}64$) compared to the best baseline method.

\item Through stratified condition-aware and data-efficiency analyses, we explicitly identify the scenarios where visual modalities excel: multimodal gains are most pronounced under cloudy conditions at $H{=}32$ and $H{=}64$. Furthermore, in few-shot settings utilizing only $5\%$ to $10\%$ of training data, \method~exhibits remarkable robustness, often achieving larger relative improvements over pure time-series models than in full-data regimes.

\end{enumerate}

The remainder of this paper is organized as follows: Section~\ref{sec:related} reviews the related work, followed by a detailed description of the proposed architecture in Section~\ref{sec:method}. Section~\ref{sec:experiments} presents the datasets, baselines, and experimental results, while Section~\ref{sec:conclusion} provides concluding remarks.

%% file: sections/related_work.tex
\section{Related work}
\label{sec:related}

\subsection{Time-series forecasting for solar power and irradiance}

Classical unimodal baselines include smart persistence, statistical methods, and predictors augmented by numerical weather prediction (NWP) or physical constraints~\cite{pedro2012assessment, inman2013solar}. While these approaches serve as robust benchmarks, particularly in data-scarce regimes, they lack the capacity to capture forward-looking cloud kinematics, relying entirely on the dynamics implicit within recent historical sequences.

To capture complex, nonlinear temporal dependencies from numerical histories, deep learning architectures have evolved significantly. Early approaches utilized recurrent networks, notably Long Short-Term Memory (LSTM)~\cite{hochreiter1997lstm}, and temporal convolutional networks~\cite{bai2018tcn}. The field subsequently shifted toward transformer-based forecasters, introducing architectures such as Informer~\cite{zhou2021informer}, Autoformer~\cite{wu2021autoformer}, and PatchTST~\cite{patchtst2023}, which leverages channel-independent patching. Concurrently, linear baselines like DLinear~\cite{zeng2023transformers} have proven highly competitive in scenarios where complex attention mechanisms are unwarranted. However, despite their advanced capacity for numerical representation, these unimodal series encoders intrinsically lack the synchronized visual context necessary to preempt abrupt meteorological shifts at the token level.

Building upon these deep learning architectures, large language models (LLMs) have recently been adapted for time-series forecasting. By leveraging frozen pretrained transformers, these approaches transfer self-attention mechanisms to sequential data with minimal task-specific adaptation~\cite{zhou2023onefitsall}. Subsequent researchers have introduced diverse alignment strategies: TEST aligns time-series embeddings with text prototypes~\cite{sun2023test}, TEMPO integrates seasonal-trend decomposition with prompt-based GPT adaptation~\cite{cao2024tempo}, $S^2$IP-LLM utilizes semantic-space anchors as prompts~\cite{pan2024s2ip}, and NNCL-TLLM constructs time-series-compatible text prototypes~\cite{bogahawatte2024nncl}. Notably, Time-LLM establishes a representative reprogramming recipe by mapping patched series into a frozen language backbone via cross-attention and prefix prompting~\cite{jin2024timellm}.This specific reprogramming framework has already been successfully translated to distributed PV power forecasting~\cite{lin2025timellm,fan2025solarllm}, reflecting a broader trend of LLM adoption across renewable energy systems for tasks ranging from forecasting to control and fault diagnosis~\cite{chandana2026llms}. While several of these advanced LLM forecasters are technically multimodal, they strictly treat the auxiliary modality as text. For instance, GPT4MTS employs textual context as soft prompts alongside numerical patches~\cite{jia2024gpt4mts}. 
A parallel line of work explores time-series foundation models, such as TimesFM~\cite{das2023decoder} and Chronos~\cite{ansari2024chronos}, which are pretrained for zero-shot forecasting and have recently been extended to support multivariate targets and numerical covariates. However, these models remain series-native and do not accommodate sky images as input. Moreover, their public ecosystem of pretrained checkpoints and task-specific recipes is less mature than that of LLMs. Ultimately, both frozen-LLM pipelines and time-series foundation models consistently omit synchronized sky imagery, leaving a critical visual gap in existing forecasting architectures.

\subsection{Sky-image and multimodal solar forecasting}

Ground-based sky cameras and geostationary satellites capture critical visual context regarding cloud kinematics that purely numerical histories cannot encode~\cite{lin2023intra,paletta2023advances,nie2024opensource}. The growing availability of public multimodal datasets now facilitates rigorous cross-site evaluation of these visual pipelines.

Purely vision-based forecasters map camera sequences directly to future irradiance or PV output. Early implementations relied on Convolutional Neural Networks (CNNs) processing single frames or short sequences~\cite{paletta2020cnn,feng2022cnn}. Subsequent advancements introduced CNN-LSTM architectures to capture extended temporal contexts~\cite{siddiqui2019skyvideos}, as well as spatiotemporal models like ECLIPSE, which explicitly track cloud motion to minimize forecast lag~\cite{paletta2022eclipse}. More recently, generative video models such as SkyGPT have been employed to synthesize future sky frames for probabilistic predictions~\cite{nie2024skygpt}. 

A parallel line of research focuses on intra-network multimodal fusion, combining sky imagery with historical numerical series. For instance, SUNSET concatenates downsampled sky frames with lagged PV measurements within a specialized CNN~\cite{sun2019sunset}, while follow-up studies have comprehensively evaluated various fusion operators for these heterogeneous inputs~\cite{venugopal2019cnn}. While these architectures successfully leverage visual motion as the primary physical driver of intra-hour variability, integrating them with an LLM-based framework is outside their design scope.

Extending this paradigm to a regional scale, satellite-based methods forecast nonlinear cloud dynamics from geostationary sequences, mapping the predicted states to PV output~\cite{si2022satellitepv,perera2026cloud}, or directly nowcasting surface solar radiation~\cite{cui2024solarnowcast}. Comprehensive frameworks like Omnivision further unify ground-based images, satellite observations, and numerical logs into a multi-view architecture~\cite{paletta2023omnivision}.

Recently, foundation models have also been explored for this domain. For instance, PV-VLM encodes sky images using a pretrained vision-language model (VLM) and fuses these embeddings with a PatchTST-style temporal encoder via late cross-modal attention~\cite{lin2025pvvlm}. In this design, the VLM operates as a semantic feature extractor, and fusion occurs after independent unimodal representations are fully formed. In contrast, \method~employs a lightweight CNN to encode sky frames, temporally aligns the resulting visual tokens with Time-LLM series patches, and applies bidirectional multimodal fusion prior to reprogramming the unified representation into a frozen language backbone. This architecture maintains a shared temporal index between modalities, ensuring cloud context is explicitly attached to the corresponding numerical look-back segments. Furthermore, rather than relying on plain concatenation which merely juxtaposes token streams without explicit cross-modal interaction, bidirectional multimodal fusion allows sky-image tokens to scale and shift series patches, and vice versa, before the modulated streams are concatenated, dimensionally reduced, and mapped.

%% file: sections/method.tex
\section{Method}
\label{sec:method}

\subsection{Problem formulation}

Let $x_{1:T} = [x_1, x_2, \dots, x_T] \in \mathbb{R}^{T}$ denote a univariate historical sequence of photovoltaic (PV) power or global horizontal irradiance (GHI). Let $I_{1:T} \in \mathbb{R}^{T \times C \times H' \times W'}$ denote the strictly synchronized sky-image sequence, where $C{=}3$ for RGB channels and each frame is resized to a spatial resolution of $H'{\times}W' = 64{\times}64$. Our goal is to forecast the future PV or GHI trajectory directly from the joint historical observations $(x_{1:T}, I_{1:T})$.

Given a historical look-back window of length $T$ and a target forecast horizon $H$, we adopt a symmetric setting where $T{=}H$ across varied horizons $H\in\{16,32,64\}$. The multi-step forecasting objective is formulated as:
\begin{equation}
\hat{y}_{T+1:T+H} = f_\theta(x_{1:T}, I_{1:T}),
\end{equation}
where $\hat{y}_{T+1:T+H} \in \mathbb{R}^{H}$ represents the predicted numerical sequence, and $f_\theta$ denotes the forecasting model parameterized by $\theta$. Finally, to ensure stable optimization and prevent data leakage, the target variables are standardized using a scaler fitted exclusively on the training split.

\subsection{Architecture: \method}

\input{figures/architecture}
Figure~\ref{fig:architecture} illustrates the overall architecture of \method, which can be conceptually divided into three main stages. 
First, the dual modality encoders process the historical solar time-series and synchronized sky image sequences into temporally aligned patch embeddings (detailed in Section~\ref{subsec:encoding}). 
Second, a Bidirectional Multimodal Fusion module explicitly models cross-modal interactions by mutually modulating the token streams (Section~\ref{subsec:fusion}). 
Finally, the fused multimodal representation undergoes time-series reprogramming (Section~\ref{subsec:prelim-ts}) and is concatenated with prompt metadata, before being processed by a frozen language backbone to yield the final forecast. Importantly, the core LLM components remain strictly frozen to preserve data-efficient adaptation, as denoted by the fire and snowflake icons.

\subsection{Preliminaries}
\label{subsec:prelim-ts}
We adopt the time-series reprogramming framework introduced in Time-LLM as the base framework for our work~\cite{jin2024timellm}. 
The normalized historical series is first segmented into overlapping patches of length $P$ with stride $S$ (configured as $P{=}4$ and $S{=}1$ in our main experiments). 
These patches are linearly projected to yield a time-series token sequence $z^{\mathrm{ts}}\in\mathbb{R}^{N\times d}$, where $N$ denotes the number of patches and $d$ is the embedding dimension. 

To bridge the modality gap between the time series patches and the LLM, the reprogramming module aligns $z^{\mathrm{ts}}$ with the language embedding space via a multi-head cross-attention mechanism. Specifically, the time-series tokens act as queries, while a set of text prototypes derived from the frozen LLM's vocabulary embedding matrix serve as keys and values. 
Following the standard Time-LLM protocol, statistical prompt tokens are constructed from the global numerical context and prepended to the aligned sequence as a Prompt-as-Prefix. 

These integrated tokens are then processed by the frozen LLM backbone. Finally, to produce the multi-step prediction, the output of the hidden states of the LLM is flattened and passed through a linear projection layer to generate the final numerical forecast $\hat{y} \in \mathbb{R}^H$.

\subsection{Image encoding and temporal alignment}
\label{subsec:encoding}
To extract spatial visual descriptors from the sky frames, we employ a lightweight CNN image encoder based on the architecture introduced in SUNSET~\cite{sun2019sunset}, which has been proven effective for sky-image feature extraction. As an alternative, we further evaluate a heavier ViT-small encoder in our ablation studies (Section~\ref{sec:experiments}).  Given the sequential image input $I_{1:T}$, the CNN encoder produces a sequence of high-dimensional feature maps.

Crucially, to synchronize the visual modality with the numerical time-series patches, these image features are temporally aggregated (e.g., via pooling over the corresponding patch windows of length $P$ and stride $S$) to match the patch count $N$. 
Subsequently, a projection Multi-Layer Perceptron (MLP) maps these aggregated visual descriptors to the identical token dimension $d$ utilized in the time-series path. 
This yields the visual token sequence $z^{\mathrm{img}}\in\mathbb{R}^{N\times d}$, which maintains a strict temporal correspondence with the patch indices $z^{\mathrm{ts}}$. Establishing this aligned, shared temporal index allows \method~to perform precise token-level multimodal conditioning before the unified representation enters the reprogramming module.

\subsection{Bidirectional multimodal fusion}
\label{subsec:fusion}

A critical architectural decision is the integration point of visual features.
If fusion occurs after reprogramming, visual cues fail to dynamically contextualize the
numerical tokens prior to their mapping into the language space.
\method~therefore introduces a Bidirectional Multimodal Fusion module before reprogramming, operating directly on the time-series patch pathway rather than acting as a replacement backbone. 
To enable mutual cross-modal conditioning, this module employs a bidirectional affine modulation mechanism inspired by feature-wise linear modulation (FiLM)~\cite{perez2018film}.

Given aligned tokens $z^{\mathrm{ts}}$ and $z^{\mathrm{img}}$, two lightweight
parameter networks, $\phi_t$ and $\phi_v$, produce affine modulation parameters:
\begin{align}
\left(g^{\mathrm{img}}, b^{\mathrm{img}}\right) &= \mathrm{split}\!\left(\phi_t(z^{\mathrm{ts}})\right), \\
\left(g^{\mathrm{ts}}, b^{\mathrm{ts}}\right) &= \mathrm{split}\!\left(\phi_v(z^{\mathrm{img}})\right).
\end{align}
The two branches are then modulated in opposite directions:
\begin{align}
\tilde{z}^{\mathrm{img}} &= g^{\mathrm{img}}\odot z^{\mathrm{img}} + b^{\mathrm{img}}, \\
\tilde{z}^{\mathrm{ts}}  &= g^{\mathrm{ts}}\odot z^{\mathrm{ts}} + b^{\mathrm{ts}}.
\end{align}
Following a non-linear processing step on each branch, a scaled residual connection is applied before fusion:
\begin{align}
\hat{z}^{\mathrm{img}} &= \sigma\!\left(\tilde{z}^{\mathrm{img}}\right) + \alpha z^{\mathrm{img}}, \\
\hat{z}^{\mathrm{ts}}  &= \sigma\!\left(\tilde{z}^{\mathrm{ts}}\right) + \alpha z^{\mathrm{ts}},
\end{align}
where $\sigma(\cdot)$ denotes a non-linear activation function, specifically the Gaussian Error Linear Unit (GELU)~\cite{hendrycks2016gaussian}, and $\alpha$ is a small scaling factor ($0.1$ in our setup).
The modulated streams are subsequently concatenated and dimensionally reduced back to $d$ via a linear projection $\psi$:
\begin{equation}
z^{\mathrm{fuse}} = \psi\!\left([\hat{z}^{\mathrm{ts}};\hat{z}^{\mathrm{img}}]\right),
\end{equation}
followed by Layer Normalization.
This bidirectional pre-reprogramming design ensures that each modality provides
contextual guidance to the other before the unified representation is projected into the
frozen language space. We compare this against a standard concatenation baseline in Section~\ref{sec:experiments}.

\subsection{Training objective}
The models are optimized to minimize the mean squared error (MSE) over the forecast horizon:
\begin{equation}
\mathcal{L} = \frac{1}{H}\sum_{h=1}^{H}\bigl(\hat{y}_{T+h}-y_{T+h}\bigr)^2.
\end{equation}
We use the Adam optimizer for up to $10$ epochs, incorporating early stopping based on validation performance. All reported testing metrics are evaluated using the best validation checkpoint under this protocol.

%% file: figures/architecture.tex
\begin{figure*}[t]
\centering
\includegraphics[width=\textwidth]{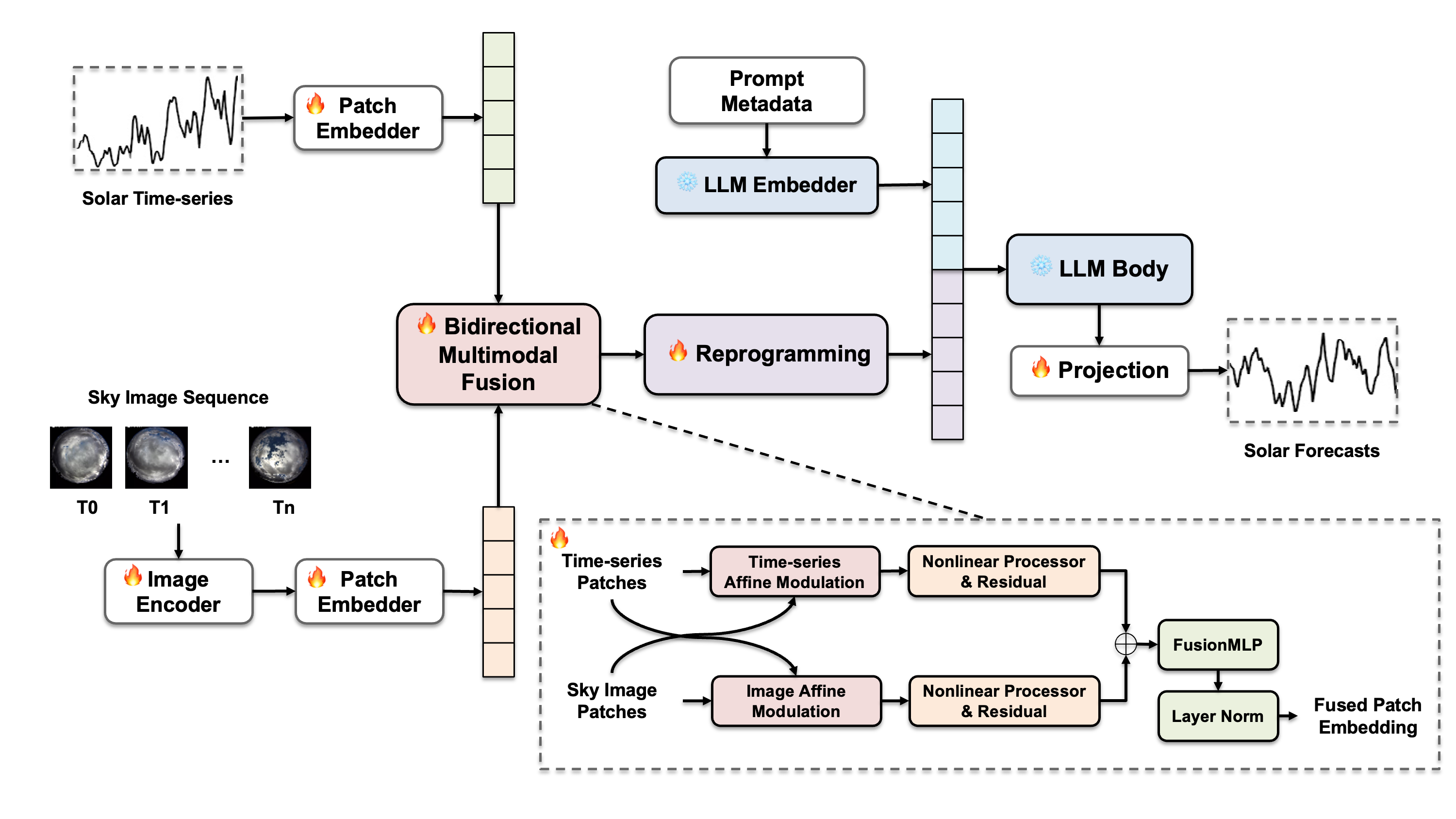}
\caption{The overall architecture of \method{}. Historical solar time-series and aligned sky images are encoded into patch tokens and integrated via a bidirectional multimodal fusion module. The fused representations then undergo patch reprogramming, followed by a frozen LLM backbone and a linear projection head to generate the final forecast. The fire and snowflake icons denote trainable and frozen components, respectively.}
\label{fig:architecture}
\end{figure*}

%% file: sections/experiments.tex
\section{Experiments}
\label{sec:experiments}

\subsection{Datasets and baselines}
We compare our method using two open-source solar datasets: SIRTA ~\cite{haeffelin2005sirta} and SKIPP'D ~\cite{nie2023skippd}.

\paragraph{SIRTA}
SIRTA provides global horizontal irradiance (GHI) series with strictly synchronized sky images from Palaiseau, France~\cite{haeffelin2005sirta}. The sampling interval is 2 min/step. The prediction horizons $H\in\{16,32,64\}$ correspond to 32, 64, and 128 minutes, respectively. For this dataset, we employ a strict chronological split: data from 2017 to 2018 is utilized for training, January to June 2019 for validation, and July to December 2019 for testing. These three periods are contiguous and mutually exclusive.

\paragraph{SKIPP'D}
SKIPP'D provides photovoltaic (PV) power series synchronized with ground-based sky images collected at the Stanford University campus in California~\cite{nie2023skippd}. The sampling interval is 1 min/step, horizons $H\in\{16,32,64\}$ corresponding to 16, 32, and 64 minutes, respectively. We adopt the exact day-level hold-out configuration established in the original dataset paper, retaining a fixed 20-day test set. We then randomly sample 100 days for validation, with the remainder serving as training data. These subsets are strictly disjoint and uniformly mixed across 2017–2019. For this dataset, we report normalized MSE and MAE with sunny/cloudy stratification following the fixed day-level partition provided by the dataset authors.

\paragraph{Common protocols}
Across both datasets, all input sky images are resized to a spatial resolution of $64\times64$. All rolling sequences are strictly constrained within a single calendar day. Target variables are standardized using scalers fitted exclusively on the training splits. Finally, for few-shot experiments, we subsample only the training days, ensuring the validation and test sets remain completely intact for rigorous evaluation.

\paragraph{Baselines}
To evaluate our proposed method, we benchmark it against a selection of prominent forecasting methods. 
Time-LLM~\cite{jin2024timellm} serves as the primary unimodal (time-series only) LLM reference, reprogramming patched numerical tokens into a frozen language backbone via cross-attention and prefix prompting. 
PatchTST~\cite{patchtst2023} represents a state-of-the-art purely numerical Transformer utilizing channel-independent patching, while DLinear~\cite{zeng2023transformers} provides a competitive linear baseline that decomposes the sequence into trend and seasonal components. 
Finally, the Smart Persistence Model (SPM) acts as our non-learning, physics-based benchmark. Unlike naive persistence, SPM extrapolates future values by holding the most recently observed clear-sky index constant and scaling it against the theoretical clear-sky profile~\cite{pedro2012assessment}. 

\subsection{Implementation details}
The fused tokens $z^{\mathrm{fuse}}$ are fed into the time-series reprogramming module, which maps them into the language embedding space via cross-attention over text prototypes derived from the LLM vocabulary. We employ a frozen GPT-2 backbone~\cite{radford2019language} truncated to $L{=}6$ layers with a native hidden size of $768$. The LLM output is subsequently flattened and passed through a linear projection head to yield the final $H$-step forecast.

For the pre-reprogramming architectures, the initial patch embedding token width is set to $d_{\mathrm{model}}{=}16$, with a feed-forward width of $128$. All models are trained using a batch size of $128$ and a base learning rate of $10^{-4}$. Specifically for \method, all parameters of the bidirectional multimodal fusion module utilize a $10\times$ learning rate multiplier by default to ensure stable multimodal alignment.

\subsection{Main results}
Tables~\ref{tab:sirta_main} and \ref{tab:skipp_main} present the overall normalized MSE and MAE for all evaluated methods across three prediction horizons. The best and second-best results in each column are highlighted in boldface and underlined, respectively. The bottom row quantifies the relative error reduction achieved by \method~compared to the strongest baseline, where positive values indicate a performance improvement.

On the SIRTA dataset, \method~consistently outperforms all baselines across every horizon. Specifically, when compared to the unimodal Time-LLM, \method~yields substantial relative MSE reductions of $20.7\%$, \textbf{$25.4\%$}, and $24.8\%$ across three horizons.

Similarly, on the SKIPP'D dataset, \method~achieves the lowest MSE at all prediction intervals, demonstrating improvements over Time-LLM by $7.0\%$ ($H{=}16$), $12.0\%$ ($H{=}32$), and \textbf{$14.5\%$} ($H{=}64$). While absolute prediction errors naturally increase with the horizon length across both datasets, the unimodal methods degrade significantly faster than our proposed framework. This divergence confirms that the benefits of visual sky-image conditioning persist, and often amplify, when forecasts must anticipate prolonged, cloud-driven ramp events.

\begin{table*}[t]
\centering
\caption{SIRTA results (normalized MSE/MAE). Best per column in bold,
second best underlined. Bottom row: relative gain of \method~vs.\ the best baseline.
$H$ denotes the symmetric look-back and forecast length in time steps, where each step corresponds to $2$~min.
}
\label{tab:sirta_main}
\setlength{\tabcolsep}{3.5pt}
\begin{tabular}{lcccccc}
\toprule
& \multicolumn{2}{c}{$H{=}16$} & \multicolumn{2}{c}{$H{=}32$} & \multicolumn{2}{c}{$H{=}64$} \\
\cmidrule(lr){2-3}\cmidrule(lr){4-5}\cmidrule(lr){6-7}
Method & MSE & MAE & MSE & MAE & MSE & MAE \\
\midrule
\method (ours) & \textbf{0.2200} & \textbf{0.2897} & \textbf{0.2557} & \textbf{0.3392} & \textbf{0.3688} & \textbf{0.4334} \\
Time-LLM & \underline{0.2775} & 0.3500 & \underline{0.3427} & 0.4212 & \underline{0.4903} & 0.5231 \\
PatchTST & 0.3000 & 0.3652 & 0.3763 & 0.4448 & 0.5756 & 0.5759 \\
DLinear & 0.3303 & 0.4170 & 0.3969 & 0.4877 & 0.6288 & 0.6351 \\
SPM & 0.3580 & \underline{0.3288} & 0.4014 & \underline{0.3684} & 0.5484 & \underline{0.4647} \\
\midrule
Gain vs.\ best baseline (\%)
  & $+20.7$ & $+11.9$
  & $+25.4$ & $+7.9$
  & $+24.8$ & $+6.7$ \\
\bottomrule
\end{tabular}
\end{table*}

\begin{table*}[t]
\centering
\caption{SKIPP'D results (normalized MSE/MAE). Best per column in bold,
second best underlined. Bottom row: relative gain of \method~vs.\ the best baseline.
$H$ denotes the symmetric look-back and forecast length in time steps, where each step corresponds to $1$~min.}
\label{tab:skipp_main}
\setlength{\tabcolsep}{3.5pt}
\begin{tabular}{lcccccc}
\toprule
& \multicolumn{2}{c}{$H{=}16$} & \multicolumn{2}{c}{$H{=}32$} & \multicolumn{2}{c}{$H{=}64$} \\
\cmidrule(lr){2-3}\cmidrule(lr){4-5}\cmidrule(lr){6-7}
Method & MSE & MAE & MSE & MAE & MSE & MAE \\
\midrule
\method (ours) & \textbf{0.1242} & \textbf{0.1560} & \textbf{0.1529} & \underline{0.1954} & \textbf{0.2069} & \textbf{0.2607} \\
Time-LLM & \underline{0.1335} & 0.1919 & \underline{0.1737} & 0.2213 & \underline{0.2421} & 0.2846 \\
PatchTST & 0.1499 & 0.2190 & 0.2143 & 0.2877 & 0.3713 & 0.3794 \\
DLinear & 0.1892 & 0.2946 & 0.2333 & 0.3512 & 0.3860 & 0.4947 \\
SPM & 0.1476 & \underline{0.1572} & 0.1856 & \textbf{0.1948} & 0.2813 & \underline{0.2661} \\
\midrule
Gain vs.\ best baseline (\%)
  & $+7.0$ & $+0.8$
  & $+12.0$ & $-0.3$
  & $+14.5$ & $+2.0$ \\
\bottomrule
\end{tabular}
\end{table*}

\subsection{Sunny/cloudy regime analysis}
Table~\ref{tab:skipp_regime} stratifies the evaluation metrics from the SKIPP'D dataset into sunny and cloudy regimes according to the fixed day-level partition defined by the dataset authors, for all evaluated methods. As anticipated, forecasting errors on cloudy days dominate the overall MSE, reflecting the high residual variance driven by transient cloud dynamics.

Under cloudy conditions, \method~consistently achieves the lowest MSE across all prediction horizons. When compared to the strongest unimodal baseline, Time-LLM, the relative performance gains of \method~scale significantly with the horizon length: $3.5\%$ at $H{=}16$, $10.8\%$ at $H{=}32$, and $13.8\%$ at $H{=}64$. 

Conversely, absolute forecasting errors on sunny days are inherently minimal. Under these clear-sky conditions, \method~remains highly competitive, with the physics-based SPM only marginally outperforming it at $H{=}32$. Ultimately, the crucial operational takeaway is that the integration of sky imagery becomes increasingly vital when models must anticipate prolonged, cloud-driven ramp events.

\begin{table*}[t]
\centering
\caption{SKIPP'D sunny/cloudy MSE. Best per column in bold,
second best underlined.
$H$ denotes the symmetric look-back and forecast length in time steps, where each step corresponds to $1$~min.}
\label{tab:skipp_regime}
\setlength{\tabcolsep}{3.5pt}
\begin{tabular}{lcccccc}
\toprule
& \multicolumn{2}{c}{$H{=}16$} & \multicolumn{2}{c}{$H{=}32$} & \multicolumn{2}{c}{$H{=}64$} \\
\cmidrule(lr){2-3}\cmidrule(lr){4-5}\cmidrule(lr){6-7}
Method & Sunny & Cloudy & Sunny & Cloudy & Sunny & Cloudy \\
\midrule
\method (ours)
  & \textbf{0.00051} & \textbf{0.2594}
  & \underline{0.00421} & \textbf{0.3154}
  & \textbf{0.00625} & \textbf{0.4182} \\
Time-LLM
  & 0.00987 & \underline{0.2688}
  & 0.00984 & \underline{0.3534}
  & 0.01167 & \underline{0.4853} \\
PatchTST
  & 0.01684 & 0.2954
  & 0.04287 & 0.4019
  & 0.05916 & 0.7002 \\
DLinear
  & 0.06250 & 0.3277
  & 0.10392 & 0.3749
  & 0.23743 & 0.5425 \\
SPM
  & \underline{0.00121} & 0.3076
  & \textbf{0.00400} & 0.3842
  & \underline{0.00688} & 0.5705 \\
\bottomrule
\end{tabular}
\end{table*}

\subsection{Few-shot data efficiency}
To evaluate sample efficiency, we subsample the training days to $10\%$ and $5\%$ while maintaining the full validation and test sets. Tables~\ref{tab:fewshot_sirta} and~\ref{tab:fewshot_skipp} report the overall MSE under these restricted budgets, with the corresponding $100\%$ results provided in Tables~\ref{tab:sirta_main} and~\ref{tab:skipp_main}. With the notable exception of $H{=}16$ on the SIRTA dataset, the deep-learning baselines (Time-LLM, PatchTST, and DLinear) perform worse than the physics-based SPM across all settings under limited training data, often by a substantial margin. This highlights a severe degradation in conventional deep forecasters to forecast longer horizons when trained on limited data. Notably, although Time-LLM relies on a frozen LLM backbone, its performance still declines, demonstrating that pretrained sequence knowledge alone is insufficient under severe data scarcity. Despite the broad collapse of baseline methods, \method~retains the best performance in most of the few-shot configurations. The sole exception occurs on the SKIPP'D dataset at $H{=}16$ ($5\%$ data), where SPM is ahead ($0.1476$ vs.\ $0.1559$). This pattern confirms that our model's robustness stems directly from the multimodal fusion rather than the time-series reprogramming alone. We attribute this resilience to two coupled architectural choices: (i) retaining a frozen LLM backbone to restrict task-specific training parameter growth, and (ii) injecting visual context via our lightweight bidirectional multimodal fusion module.

\begin{table*}[t]
\centering
\caption{SIRTA few-shot overall MSE (day-level train subsample, $10\%$/$5\%$ only).
Full-data ($100\%$) results are in Table~\ref{tab:sirta_main}.
Best per column in bold, second best underlined.}
\label{tab:fewshot_sirta}
\setlength{\tabcolsep}{4.0pt}
\begin{tabular}{lcccccc}
\toprule
& \multicolumn{2}{c}{$H{=}16$} & \multicolumn{2}{c}{$H{=}32$} & \multicolumn{2}{c}{$H{=}64$} \\
\cmidrule(lr){2-3}\cmidrule(lr){4-5}\cmidrule(lr){6-7}
Method & $10\%$ & $5\%$ & $10\%$ & $5\%$ & $10\%$ & $5\%$ \\
\midrule
\method (ours)
  & \textbf{0.2636} & \textbf{0.3047}
  & \textbf{0.2943} & \textbf{0.3283}
  & \textbf{0.4423} & \textbf{0.4952} \\
Time-LLM
  & \underline{0.3061} & \underline{0.3099}
  & 0.4138 & 0.4163
  & 0.7502 & 0.7527 \\
PatchTST
  & 0.3349 & 0.3381
  & 0.4573 & 0.4681
  & 0.7778 & 0.8163 \\
DLinear
  & 0.3472 & 0.3485
  & 0.4241 & 0.4275
  & 0.7477 & 0.7616 \\
SPM
  & 0.3580 & 0.3580
  & \underline{0.4014} & \underline{0.4014}
  & \underline{0.5484} & \underline{0.5484} \\
\bottomrule
\end{tabular}
\end{table*}

\begin{table*}[t]
\centering
\caption{SKIPP'D few-shot overall MSE (day-level train subsample, $10\%$/$5\%$ only).
Full-data ($100\%$) results are in Table~\ref{tab:skipp_main}.
Best per column in bold, second best underlined.}
\label{tab:fewshot_skipp}
\setlength{\tabcolsep}{4.0pt}
\begin{tabular}{lcccccc}
\toprule
& \multicolumn{2}{c}{$H{=}16$} & \multicolumn{2}{c}{$H{=}32$} & \multicolumn{2}{c}{$H{=}64$} \\
\cmidrule(lr){2-3}\cmidrule(lr){4-5}\cmidrule(lr){6-7}
Method & $10\%$ & $5\%$ & $10\%$ & $5\%$ & $10\%$ & $5\%$ \\
\midrule
\method (ours)
  & \textbf{0.1296} & \underline{0.1559}
  & \textbf{0.1639} & \textbf{0.1719}
  & \textbf{0.2269} & \textbf{0.2693} \\
Time-LLM
  & 0.1575 & 0.1633
  & 0.2392 & 0.2414
  & 0.4570 & 0.4767 \\
PatchTST
  & 0.1654 & 0.1672
  & 0.2523 & 0.2594
  & 0.4631 & 0.4920 \\
DLinear
  & 0.1992 & 0.2004
  & 0.2484 & 0.2501
  & 0.4627 & 0.4727 \\
SPM
  & \underline{0.1476} & \textbf{0.1476}
  & \underline{0.1856} & \underline{0.1856}
  & \underline{0.2813} & \underline{0.2813} \\
\bottomrule
\end{tabular}
\end{table*}

\subsection{Fusion ablation}
Table~\ref{tab:fusion_main} compares the full \method~ bidirectional multimodal fusion architecture against a standard concatenation (Concat) baseline under the established setting across $H\in\{16,32,64\}$.

On the SIRTA dataset, \method demonstrates superior performance over the Concat baseline at all three horizons. For SKIPP'D, our approach maintains its advantage at $H{=}16$ and $H{=}32$, while Concat achieves a marginally lower error at the longest horizon of $H{=}64$.

\begin{table}[t]
\centering
\caption{Fusion ablation (overall normalized MSE). \method~uses bidirectional
multimodal fusion; Concat replaces only the fusion operator.}
\label{tab:fusion_main}
\begin{tabular}{lcccccc}
\toprule
& \multicolumn{3}{c}{SIRTA} & \multicolumn{3}{c}{SKIPP'D} \\
\cmidrule(lr){2-4}\cmidrule(lr){5-7}
Method & $H{=}16$ & $H{=}32$ & $H{=}64$ & $H{=}16$ & $H{=}32$ & $H{=}64$ \\
\midrule
\method & \textbf{0.2200} & \textbf{0.2557} & \textbf{0.3688} & \textbf{0.1242} & \textbf{0.1529} & 0.2069 \\
Concat & 0.2339 & 0.2672 & 0.4017 & 0.1287 & 0.1855 & \textbf{0.1931} \\
\bottomrule
\end{tabular}
\end{table}

\subsection{Image encoder ablation}
\label{sec:encoder_abl}
To evaluate the impact of the visual backbone, we substitute the default CNN encoder with a ViT-small~\cite{dosovitskiy2020image} architecture, while keeping the bidirectional multimodal fusion module unchanged. Table~\ref{tab:encoder_main} reports the overall normalized MSE on both datasets across $H\in\{16,32,64\}$.
On the SIRTA dataset, the CNN encoder consistently outperforms the ViT across all horizons, exhibiting relative MSE advantages of $7.3\%$ ($H{=}16$), $11.6\%$ ($H{=}32$), and $4.7\%$ ($H{=}64$). Conversely, on the SKIPP'D dataset, the ViT yields slightly worse performance at $H{=}16$ ($+4.2\%$) and $H{=}32$ ($+8.8\%$), though it performs marginally better at the longest horizon, $H{=}64$ ($-3.5\%$).

Given that the CNN encoder outperforms the ViT in the majority of scenarios and its lightweight architecture inherently benefits few-shot data efficiency, we conclude that the CNN is the superior choice for our primary experiments.

\begin{table}[t]
\centering
\caption{Image encoder ablation (overall normalized MSE). CNN is the default
\method~image encoder. ViT-small replaces only the image backbone.}
\label{tab:encoder_main}
\begin{tabular}{lcccccc}
\toprule
& \multicolumn{3}{c}{SIRTA} & \multicolumn{3}{c}{SKIPP'D} \\
\cmidrule(lr){2-4}\cmidrule(lr){5-7}
Encoder & $H{=}16$ & $H{=}32$ & $H{=}64$ & $H{=}16$ & $H{=}32$ & $H{=}64$ \\
\midrule
CNN (default) & \textbf{0.2200} & \textbf{0.2557} & \textbf{0.3688} & \textbf{0.1242} & \textbf{0.1529} & 0.2069 \\
ViT-small & 0.2360 & 0.2853 & 0.3860 & 0.1294 & 0.1664 & \textbf{0.1996} \\
\bottomrule
\end{tabular}
\end{table}

%% file: sections/conclusion.tex
\section{Conclusion}
\label{sec:conclusion}

In this work, we introduced \method, a multimodal forecasting framework that combines contemporaneous sky image features into an LLM-based time-series forecaster for solar forecasting. To achieve this, we propose a lightweight bidirectional multimodal fusion module that enables explicit cross-modal interaction through token-wise affine modulation. This design effectively conditions the numerical tokens before they enter the frozen LLM backbone, allowing the visual and numerical modalities to mutually modulate one another and fostering a richer, deeply integrated cross-modal representation. Extensive experiments on two open-source solar datasets demonstrate that \method~consistently outperforms the best baseline across all tested horizons ($H\in\{16,32,64\}$), achieving peak relative MSE reductions of $25.4\%$ on SIRTA ($H{=}32$) and $14.5\%$ on SKIPP'D ($H{=}64$).

Our proposed framework excels in severe data-scarcity scenarios and at longer forecast horizons under cloudy conditions, where unimodal approaches struggle.
In few-shot scenarios using only $5\%$ to $10\%$ of available training days, conventional deep-learning baselines frequently underperform the physics-based SPM. In contrast, \method~demonstrates remarkable resilience, leveraging the combination of a frozen LLM backbone and compact cross-modal modulation to maintain superior forecasting performance. Our ablation studies further validate the advantages of the proposed bidirectional multimodal fusion module in effectively integrating these disparate modalities in most cases. 

As a primary limitation, we acknowledge that ground-based sky cameras inherently provide only localized and short-term cues regarding cloud dynamics. Consequently, future work will focus on developing a more comprehensive multimodal framework that simultaneously incorporates ground-based sky images, large-scale satellite imagery, and text-based weather forecasts. By synthesizing these diverse data sources, it is possible to deliver robust solar forecasting capabilities across both short and extended prediction horizons.